\documentclass{ecai} 

\usepackage{latexsym}
\usepackage{amssymb}
\usepackage{amsmath}
\usepackage{amsthm}
\usepackage{booktabs}
\usepackage{enumitem}
\usepackage{graphicx}
\usepackage{color}

\usepackage{fontawesome}
\usepackage{adjustbox}
\usepackage{hyperref}
\hypersetup{hidelinks}
\usepackage[ruled,linesnumbered]{algorithm2e}

\newcommand{\faY}[0]{\faStar}
\newcommand{\faH}[0]{\faStarHalfO}
\newcommand{\faN}[0]{\faTimes}

\newcommand{\BibTeX}{B\kern-.05em{\sc i\kern-.025em b}\kern-.08em\TeX}

\begin{document}


\begin{frontmatter}


\paperid{628} 


\title{Prompt Recursive Search: A Living Framework with Adaptive Growth in LLM Auto-Prompting}






\author[A]{\fnms{Xiangyu}~\snm{Zhao}\orcid{0009-0007-0903-0352}}

\author[A]{\fnms{Chengqian}~\snm{Ma}\orcid{0009-0005-9269-3880}}

\address[A]{Xiamen University}


\begin{abstract}
Large Language Models (LLMs) exhibit remarkable proficiency in addressing a diverse array of tasks within the Natural Language Processing (NLP) domain, with various prompt design strategies significantly augmenting their capabilities. However, these prompts, while beneficial, each possess inherent limitations. The primary prompt design methodologies are twofold: The first, exemplified by the Chain of Thought (CoT), involves manually crafting prompts specific to individual datasets, hence termed \textbf{Expert-Designed Prompts (EDPs)}. Once these prompts are established, they are unalterable, and their effectiveness is capped by the expertise of the human designers. When applied to LLMs, the static nature of EDPs results in a uniform approach to both simple and complex problems within the same dataset, leading to the inefficient use of tokens for straightforward issues. The second method involves prompts autonomously generated by the LLM, known as \textbf{LLM-Derived Prompts (LDPs)}, which provide tailored solutions to specific problems, mitigating the limitations of EDPs. However, LDPs may encounter a decline in performance when tackling complex problems due to the potential for error accumulation during the solution planning process. To address these challenges, we have conceived a novel \textbf{Prompt Recursive Search (PRS)} framework that leverages the LLM to generate solutions specific to the problem, thereby conserving tokens. The framework incorporates an assessment of problem complexity and an adjustable structure, ensuring a reduction in the likelihood of errors. We have substantiated the efficacy of PRS framework through extensive experiments using LLMs with different numbers of parameters across a spectrum of datasets in various domains. Compared to the CoT method, the PRS method has increased the accuracy on the BBH dataset by 8\% using Llama3-7B model, achieving a 22\% improvement.

\end{abstract}

\end{frontmatter}

\begin{table}[h]
\centering
\caption{Analysis of prompt methodologies in relation to their capabilities, utilizing the following symbols: ``\faY'' for comprehensive support, ``\faH'' for limited support, and ``\faN'' for absence of support. CE: Computational Resource Utilization Efficiency, IH: Independence from Human Expertise, SE: Supervision of Errors in the Inference Process.}
\vspace{10pt}
\begin{adjustbox}{width=0.3\textwidth} 
\setlength{\tabcolsep}{1.5pt}
\footnotesize
\begin{tabular}{llllll}
\toprule
\textbf{Scheme} & \textbf{CE} & \textbf{IH} & \textbf{SE}\\
\midrule
Expert-Designed Prompt~\citep{Wei2022ChainOT,Wang2023PlanandSolvePI,Yao2023TreeOT,besta2024aaai} &  \faN & \faN & \faH\\
LLM-Derived Prompt~\citep{Pryzant2023AutomaticPO,Levi2024IntentbasedPC,Zhou2022LargeLM} &  \faY & \faY & \faN\\
Prompt Recursive Search~ &  \faY & \faY & \faY\\
\bottomrule
\end{tabular}
\end{adjustbox}

\label{tab:schemes}
\end{table}

\section{Introduction}
\label{sec:intro}

\begin{figure*}[h!]
\centering
\includegraphics[width=\linewidth]{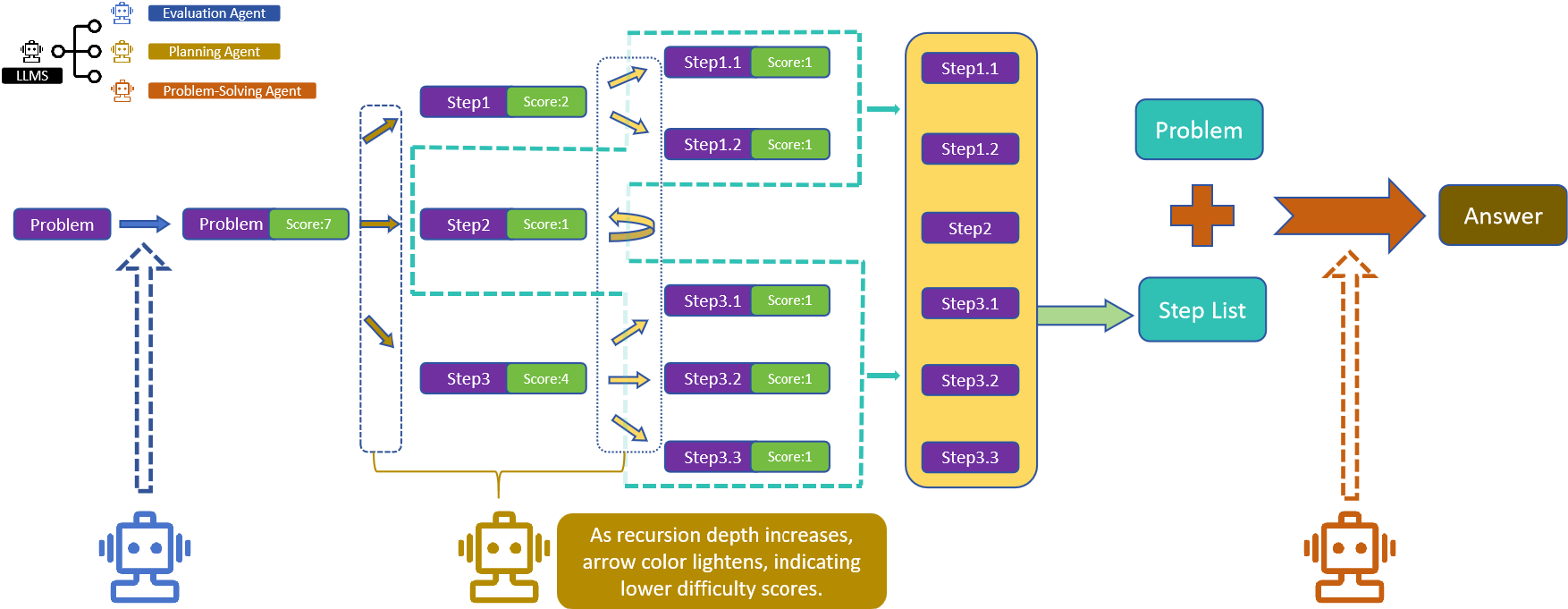}
\caption{Pipeline of our proposed method Prompt Recursive Search with complexity score limitation (PRS).}
\label{fig:pipeline}
\end{figure*}

Stem cells differentiate into other types of cells with distinct functions upon induction by signaling molecules\citep{Becker1963CytologicalDO,Siminovitch1963THEDO}. A similar process, as depicted in \ref{fig:pipeline}, is proposed by us in the use of Large Language Models (LLMs) to solve problems. Previous work\citep{Wei2022ChainOT,Kojima2022LargeLM,Zhang2022AutomaticCO,Yu2023ThoughtPA,Madaan2023SelfRefineIR} has indicated that a single interaction often fails to adequately address problems. Instead, a more effective approach is to divide the problem-solving process into multiple steps: first proposing a solution to the problem, and then executing it step by step by the LLM. Each interaction with the LLM involves input to the model and its subsequent output. A single interaction can lead to the refinement, detailing, summarization, or execution of a solution. Each interaction corresponds to the realization of a thought within the thoughts that make up the solution. The functionality of a thought evolves from a more basic state to become increasingly specific, much like how a stem cell differentiates into a cell with a particular function. Similarly, an original thought can differentiate in response to various problems, giving rise to thoughts capable of solving different issues.

Control over the evolution of thoughts traditionally involves two methods. The first method entrusts the transformation process of thoughts entirely to experienced domain experts. These experts can, based on their understanding of problems within a particular field, plan the problem-solving process in a sequential chain, addressing the problem step by step. This is the contribution of the Chain of Thought (CoT) approach\citep{Wei2022ChainOT}. Prior to this, the practice of presenting a problem to LLMs and expecting a direct answer placed excessive demands on the LLMs, yielding suboptimal results and failing to fully leverage LLM's potential. Given that there is often more than one method to solve a problem, and multiple approaches can yield answers for a specific issue, we can compare the answers derived from various methods to select the most optimal one, thus obtaining the best solution to the problem. Plan-and-Solve (PS)\citep{Wang2023PlanandSolvePI} Prompting is a concrete implementation of this line of thinking.
Building upon this, the Tree of Thought (ToT)\citep{Yao2023TreeOT} further extends this work by considering a scenario where a thought represents a critical step in the solution. If this thought is flawed, continuing to reason based on it can lead to more severe errors. Therefore, the authors of ToT propose a prompt structure that allows for a retreat from an erroneous solution path to reconsider and pursue the correct approach. This structure is hierarchical, and the feature is known as the backtracking function during the tree traversal process.
On the foundation of ToT, the Graph of Thought (GoT)~\citep{besta2024aaai} structure is even more flexible. By performing operations such as aggregation and refinement on multiple thoughts, the expert-designed prompt structure represented by GoT is highly fault-tolerant. The outcome of one thought can be verified by multiple thoughts, and the knowledge derived from multiple thoughts can be integrated into a single thought.

\begin{figure*}[t]
\centering
\includegraphics[width=\linewidth]{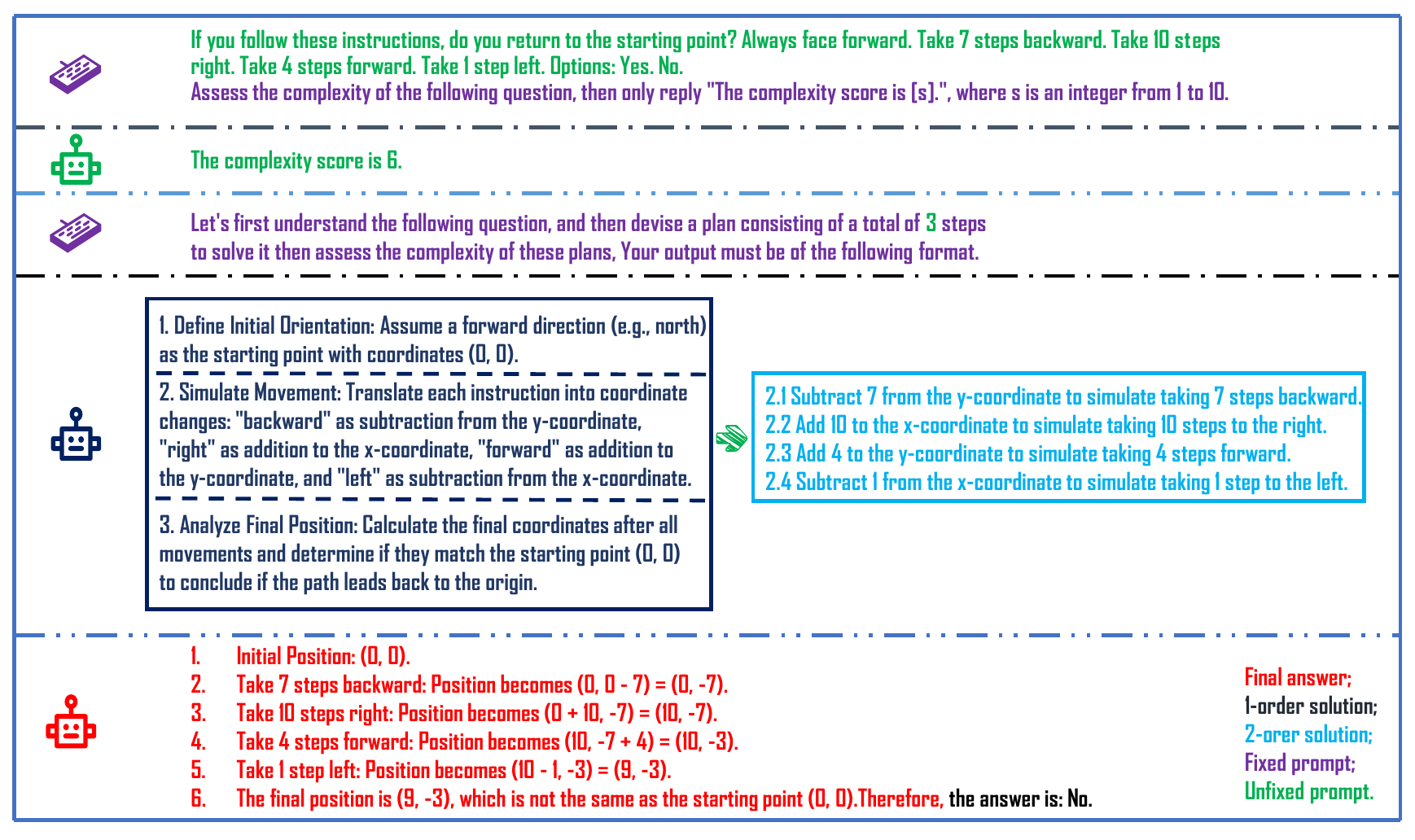}
\caption{Example of our proposed method Prompt Recursive Search with complexity score limitation (PRS).}
\label{fig:examplar}
\end{figure*}

We refer to the prompt design structures represented by CoT (Chain of Thought), ToT (Tree of Thought), and GoT (Graph of Thought) as \textbf{Expert-Designed Prompts (EDP)}: These prompt frameworks are manually crafted, thus necessitating a substantial amount of human expert experience. The approach demands a high level of expertise from the designer, who must draw on their professional experience and conduct multiple experiments to derive a prompt structure capable of addressing various scenarios within a particular domain. Given that this method accounts for a multitude of potential prompt structures, it is universally effective for all problems, which may result in a complex structure. While this approach often excels in tackling complex issues, it can be inefficient for simpler problems, as many steps within the prompt structure become redundant, leading to unnecessary computational resource wastage. Furthermore, if the human expert responsible for designing the prompt lacks sufficient proficiency or fails to consider all possible scenarios comprehensively, the static nature of the prompt, once designed, can lead to the persistent presence of any inherent flaws in the structure during each application.

To mitigate the dependency of Large Language Model (LLM) prompt design on human experience, a second approach to prompt design has been proposed. The primary issues with human-engineered prompt structures are twofold: Firstly, due to the limitations of human cognition, human experts cannot fully account for all scenarios within a domain. Secondly, once a prompt structure is designed by a human expert, it becomes immutable, which can lead to the waste of substantial computational resources when addressing simple problems, as the comprehensively complex structure, initially a strength, becomes redundant.
However, entrusting the task of prompt structure design to an LLM addresses these concerns. Given the vast training corpora of large language models, their breadth of knowledge can surpass that of individual human experts. Additionally, the prompt framework design functionality provided by LLMs can commence upon the presentation of a specific problem, thereby being tailored to that particular issue. For problems that are straightforward and can be resolved in a single step, the LLM-generated prompt structure will not incur the unnecessary computational resource expenditure associated with the redundant steps found in human expert-designed prompt structures.

In the work of Automatic Prompt Engineering (APE)\citep{Zhou2022LargeLM}, the Large Language Model (LLM) initially proposes a set of prompts that encompass solutions to specific problems. The quality of these prompts varies, leading to a selection process where lower-quality prompts are discarded, and higher-quality ones are retained. Then LLM can use the higher-quality prompts as references to generate similar ones, enhancing the diversity of the prompt collection. This iterative process continues until satisfactory prompts are obtained. During the prompt generation by the LLM, the model can act not only as a problem solver providing solutions to domain-specific questions but also as a question generator, posing some extreme or marginal questions within a domain. These questions can be utilized to optimize the prompts, thereby improving their quality.
Building on this approach, Intent-based Prompt Calibration (IPC)\citep{Levi2024IntentbasedPC} optimizes the prompts using these edge-case questions after designing the prompts and related questions, ensuring the quality of the prompts by leveraging these domain-marginal issues. Following the proposal of prompts by the LLM, to ensure their quality, we can employ methods mentioned in the first two approaches: evaluating and discarding prompts with lower quality, or constraining them with edge-case questions. Additionally, we can utilize the Automatic Prompt Optimization\citep{Pryzant2023AutomaticPO} method: composing all possible prompts into a prompt space, using a prompt generated by the LLM as the starting point within this space, and optimizing the prompt based on its evaluation from LLM as the textual gradient. Through repeated iterations, the optimal solution can be found within the prompt space, employing a textual gradient descent method to identify the most ideal solution. The solutions proposed by the LLM, in conjunction with the problem context or textual gradient, often achieve effects comparable to those of human-designed prompts, which we term as \textbf{LLM-Derived Prompts (LDP)}.

However, when dealing with excessively complex problems, such as large-scale planning issues, the complete delegation of the planning of solutions to the LLM can become problematic. With numerous steps involved, there is a high likelihood of error propagation and accumulation. A minor error at any stage can be rapidly magnified across multiple steps, leading to suboptimal performance by the LLM.

Expert-Designed Prompts (EDPs) often necessitate a more complex structure to address all scenarios within a dataset, which can result in redundant steps when EDPs are applied to solve simpler problems within that dataset. On the other hand, LLM-Derived Prompts (LDPs) delegate the entire task of prompt design to the Large Language Model (LLM). During the process of generating solutions to problems, the LLM is highly susceptible to the accumulation of errors, which significantly increases the difficulty of accurately generating solutions for complex tasks.

We have identified that the characteristics of Expert-Designed Prompts (EDPs) and LLM-Derived Prompts (LDPs) are, to a certain extent, complementary. Consequently, after thorough consideration of the advantages and disadvantages of both EDPs and LDPs, we propose an entirely new \textbf{Prompt Recursive Search(PRS)} framework for prompt design which is depicted in \ref{fig:pipeline}. This framework enables Large Language Models (LLMs) to avoid overly complex planning when tackling simplistic problems, while also allowing LLMs to undertake a portion of the prompt design work, thus reducing the high dependency on human experts. Drawing inspiration from the differentiation process of human stem cells: when the body, prompted by signaling molecules, requires cells to perform a specific task, stem cells differentiate into new cells with distinct characteristics to address issues that other cells cannot. Our prompt framework, when confronted with a new problem, first evaluates the problem and has the LLM assign a complexity score. If the problem's complexity is deemed too high, we break down the problem into multiple steps for resolution. The simplified problems, post-dissection, become new targets, for which the LLM provides solutions. This is akin to allowing the LLM to complete a differentiation design process for the prompt, resulting in prompts with specific functionalities. During the differentiation steps, we recursively apply this differentiation function until the problem has been sufficiently broken down into simple components.

By employing LLMs with varying numbers of parameters, we compared the PRS method with traditional prompt design approaches on the BBH dataset, which spans multiple domains, thereby validating the effectiveness of the PRS method. Through this paper, we have made the following contributions:

\begin{itemize}
\setlength{\itemsep}{0pt}
\setlength{\parsep}{0pt}
\setlength{\parskip}{0pt}
    \item We have categorized existing prompts based on their source of acquisition into two types: Expert-Designed Prompts (EDP) and LLM-Derived Prompts (LDP);
    \item By introducing biological principles, we have proposed an automated framework for addressing complex problems that integrates the advantages of both EDP and LDP.
    \item Through validation on the BBH\citep{Suzgun2022ChallengingBT} dataset across multiple domains using models with parameters spanning different orders of magnitude, we have confirmed the effectiveness of this method. 
\end{itemize}

\section{Related Work}
\label{sec:Related Work}

\subsection{LLM-Derived Prompts}

Large Language Models (LLMs) are an advanced tool in the field of Natural Language Processing (NLP) technology\citep{Zhao2023ExplainabilityFL,Zhao2023ASO}. They are based on deep learning architectures, such as Transformers\citep{Vaswani2017AttentionIA}, and have learned complex patterns of language through extensive training on vast amounts of text data. These models are not only capable of understanding the nuances of language but also generating coherent and grammatically correct text, thereby performing well on a variety of language tasks. They can be used in a wide range of applications, including chat-bots\citep{Peng2024AIRO}, recommendation systems\citep{Liu2023AFL}, content creation aids\citep{Cao2023ACS}, and educational tools\citep{Orenstrakh2023DetectingLT, Mvondo2023GenerativeCA}. However, LLMs may also replicate and amplify biases present in their training data, so ethical\citep{Cabrera2023EthicalDM, Goetz2023UnreliableLB, Bang2023ExaminationOE, Mvondo2023GenerativeCA} and bias\citep{Yeh2023EvaluatingIL, Oketunji2023LargeLM} considerations must be taken into account in their design and use. Notable examples of LLMs include the BERT\citep{Devlin2019BERTPO} and GPT\citep{Radford2018ImprovingLU} models, which have achieved significant success in tasks involving natural language understanding\citep{Radford2018ImprovingLU, Wang2018GLUEAM} and generation\citep{Lewis2019BARTDS, Ji2022SurveyOH}.

\subsection{Expert-Designed Prompts}
\subsubsection{Chain of Thought(CoT)}
This technology is designed to enhance the ability of Large Language Models (LLMs) to handle complex issues. It simulates the continuous thought process of human problem-solving by embedding a series of logical reasoning steps within the input prompts. These steps progressively guide the model to the solution of the problem, aiding in the processing of tasks that require continuous logic or mathematical computation. By demonstrating examples of problem-solving, the accuracy and logicality of the model's output can be improved, even without specific task training.
\subsubsection{Plan-and-Solve (PS) Prompting}
Compared with Chain of Thought (CoT), Plan-and-Solve Prompting significantly improves the performance of Large Language Models (LLMs) in multi-step reasoning tasks by guiding the models to formulate plans for solutions and then execute them. It has demonstrated superior performance over Zero-shot-CoT thinking across multiple datasets and is comparable to CoT methods that require manual examples, showcasing the potential to stimulate the reasoning capabilities of LLMs without the need for manual examples.
\subsubsection{Tree of Thoughts}
The Tree of Thought (ToT) framework aims to enhance the capabilities of Large Language Models (LLMs) in problem-solving. It allows the model to explore various reasoning paths and to self-assess its decisions, leading to more deliberate choices. This approach views problem-solving as a search process within a "thought tree," where each "thought" is a coherent text block that represents a step towards the solution.
\subsubsection{Graph of Thoughts}
Graph of Thoughts (GoT) significantly enhances the performance of Large Language Models (LLMs) in complex task processing by conceptualizing the reasoning process as a graph structure composed of nodes, which represent the model's "thoughts," and edges, which denote the connections between these thoughts. Not only does GoT improve the quality of task execution, with a 62\% increase in efficiency over existing technologies in sorting tasks, but it also achieves a reduction in costs by more than 31\%. The design of GoT greatly facilitates the approximation of LLMs' reasoning capabilities to human thought patterns.


\subsubsection{Automatic Prompt Engineer(APE)}
The manual design of prompts necessitates a high level of expertise from the designer and is constrained by the relatively limited knowledge base and subjective factors of human experts compared to Large Language Models (LLMs). Handcrafted prompts are subject to numerous limitations stemming from human factors, which LLMs are not bound by. Therefore, entrusting the task of prompt design to an LLM can both resolve these constraints and eliminate the inconvenience of manual prompt design, thereby automating the use of LLMs. Initially, the LLM is tasked with generating a set of prompts, which are then not utilized in their entirety. Instead, these prompts are evaluated and the highest-quality ones are selected. Based on the highest-quality prompts, a variety of similar prompts are generated to enhance their diversity. 

\subsubsection{Intent-based Prompt Calibration(IPC)}
Large Language Models (LLMs) are highly sensitive to the input content, where even seemingly irrelevant random noise can significantly alter the LLM's output. The root cause of this sensitivity is a lack of an accurate and comprehensive understanding of the problem. To ensure that LLMs consider the problem thoroughly, based on the user's requirements and the initial prompt, new prompts and data samples are iteratively generated. The prompts are then optimized using edge data, allowing the LLM to fully contemplate the problem. This operation of generating data samples opens up new avenues for prompt design: not only can we have LLMs generate answers for us, but we can also have LLMs generate questions that can, in turn, lead to better answers.

\subsubsection{Automatic Prompt Optimization}
After the generation of prompts by a LLM, it is often necessary to refine them. This refinement can be accomplished through simple ranking and selection, known as Approach by APE, or by making modifications to the prompt based on the data provided by the LLM, referred to as IPC. However, when the objective is to identify the most appropriate prompt within a dense space of prompts, gradient descent emerges as a more rational method. The process begins with the LLM generating a response based on an initial prompt. Subsequently, the LLM analyzes the response in conjunction with the prompt to produce Textual Gradients. These Textual Gradients are then integrated to formulate a new prompt. This cycle of generation and refinement is repeated iteratively until the optimal prompt is achieved. 

\section{Methodology}
\label{sec:Methodology}

Our approach is founded upon the principles of the EDP and the LDP, taking into account that while EDP is adept at addressing straightforward problems with a comprehensive framework designed to encompass all scenarios of such issues, it may necessitate superfluous token expenditure when applied to simpler problems. On the other hand, LDP can lead to the accumulation of errors; if an LLM is tasked with designing prompts without constraints or corrections, it must adhere to a process that involves planning a solution and then executing it. Errors introduced during the planning phase can propagate through to subsequent steps, potentially undermining the problem-solving process. To mitigate the costs associated with manually designed prompts, we delegate the responsibility of devising solutions to the LLM itself. The LLM will design tailored solutions for each problem, thereby avoiding unnecessary token usage. Furthermore, to prevent the accumulation of errors during the LLM's prompt design phase, we employ complexity detection methods and procedural planning techniques to constrain the LLM and validate its solution planning. Our method is showed in \ref{alg:design_solution} and \ref{alg:PRS}.

\subsection{Evaluating the Complexity of a Problem}
Before devising solutions for a problem, we first assess its complexity to facilitate the design of more comprehensive solutions for more complex issues and more straightforward solutions for simpler ones. The complexity level can provide effective guidance for formulating our solutions. In the process of evaluating the complexity of a problem by a Large Language Model (LLM), we consider the characteristics of textual descriptions of problems: directly describing the complexity using natural language can lead to an inability to distinctly differentiate between problems based on this metric. Therefore, we opt to use numerical values instead of the original complexity descriptions. We categorize the complexity of problems into ten levels, represented by ten integers ranging from 1 to 10.

\subsection{Planning Solution for a Problem}

Upon determining the complexity of a problem, our focus shifts to designing solutions based on that complexity. In practice, the aspect of a solution that is directly related to complexity is the number of steps involved. Through extensive experimentation, we have correlated the complexity levels proposed by the Large Language Model (LLM) for problems with the steps required in their solutions. Our findings indicate a linear relationship: the number of steps to resolve a problem typically ranges between one and five, and the complexity level of a problem, when divided by two, often yields the number of steps needed for its resolution. This macroscopic statistical pattern allows us to seamlessly incorporate the initial complexity assessment into the design of solutions, thereby specifying the number of steps required for resolution.

\begin{algorithm}[t]
\small
\caption{Design Solution}\label{alg:design_solution}
\KwIn{Problem $\mathbf{P}$, Steps $\mathbf{Steps}$}
\KwOut{Solution $\mathbf{Solution}$}

$\mathbf{Solution} \gets [\ ]$

$\mathbf{Initial\_Steps} \gets \text{Get\_Initial\_Solution}(\mathbf{P}, \mathbf{Steps})$

\ForEach{$\mathbf{Step}$ \textbf{in} $\mathbf{Initial\_Steps}$}{
    $\mathbf{Complexity} \gets \text{Evaluate\_Complexity}(\mathbf{Step})$
    
    \eIf{$\mathbf{Complexity} > \text{Threshold}$}{
        $\mathbf{New\_Steps} \gets \text{Calculate\_Steps}(\mathbf{Complexity})$
        
        $\mathbf{Solution} \gets \mathbf{Solution} + \text{Design\_Solution}(\mathbf{Step}, \mathbf{New\_Steps})$
    }
    {
        $\mathbf{Solution} \gets \mathbf{Solution} + [\mathbf{Step}]$
    }
}

\Return $\mathbf{Solution}$
\end{algorithm}

\begin{algorithm}[t]
\small
\caption{PRS Pipeline}\label{alg:PRS}
\KwIn{Problem description $\mathbf{P}_{desc}$}
\KwOut{Answer $\mathbf{Ans}$}

\tcp{Evaluate Complexity}
$\mathbf{C} \gets \text{Evaluate\_Complexity}(\mathbf{P}_{desc})$

\tcp{Calculate Steps}
$\mathbf{Steps} \gets \text{Calculate\_Steps}(\mathbf{C})$

\tcp{Design Solution}
$\mathbf{Sol} \gets \text{Design\_Solution}(\mathbf{P}_{desc}, \mathbf{Steps})$

\tcp{Get Answer}
$\mathbf{Ans} \gets \text{Get\_Answer}(\mathbf{P}_{desc}, \mathbf{Sol})$ 

\Return $\mathbf{Ans}$
\end{algorithm}
\subsection{Recursively Proposing Solutions to Problems}
Following the first two steps, we have established the necessary steps for a solution, which are determined based on the complexity of the problem. This approach, underpinned by our discovered relationship between complexity levels and solution steps, mitigates the accumulation of errors during the problem formulation by the LLM. It addresses the shortcomings of the LLM-Derived Prompt (LDP) while fully leveraging its strengths in devising tailored solutions for specific problems, thereby also resolving the issues associated with the Expert-Designed Prompt (EDP).

\subsection{Obtaining a complete solution}
Subsequently, we can employ the identified solution steps to propose resolutions. Recognizing that providing detailed steps for complex problems in a single response demands a high level of proficiency from the LLM, we adopt a recursive strategy for the more intricate steps of the solution. Each complex step is treated as a new problem to be broken down, and a more refined solution is proposed, prompting the LLM to elaborate on the specifics of the solution. This recursive process continues until all solutions are simplified to the point where they can be resolved in a single step. By transforming complex problems into simpler ones, we prevent the LLM from being overwhelmed by complexity and ensure the provision of accurate solutions.

\subsection{Obtaining an Answer according to the solution}
After obtaining a solution through the preceding plan, we can then apply this solution to address the problem. Additionally, we can impose constraints on the format of the answers outputted by the Large Language Model (LLM) with respect to specific datasets, ensuring the production of coherent and appropriate responses.

\section{Experiments}
\label{sec:Experiments}

\begin{figure*}[t]
\centering
\includegraphics[width=\linewidth]{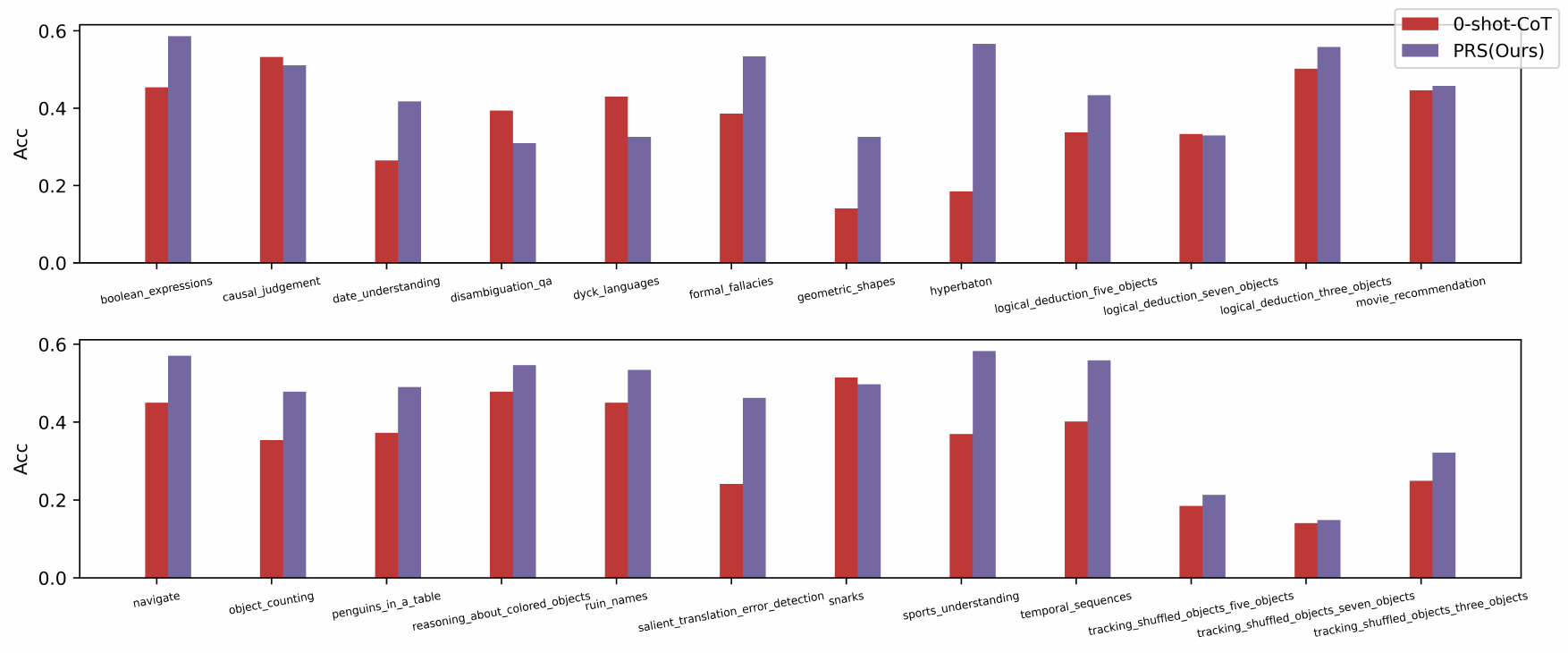}
\caption{Comparison between PRS (ours) and 0-shot-cot on the BBH dataset.}
\label{fig:result}
\end{figure*}

\subsection{Model Selection}
In the process of model selection, considering that Large Language Models (LLMs) with a substantial number of parameters inherently possess higher performance, the advent of prompt design methodologies has been predominantly aimed at enhancing the capabilities of LLMs with moderate to smaller parameter counts. Moreover, the efficacy of prompt design methods is more pronounced in LLMs with smaller parameter volumes. We can enhance the performance of LLMs by designing prompts, fully tapping into the potential of LLMs; however, LLMs with a larger number of parameters already possess relatively good capabilities, and even with the use of prompt techniques, the improvement in LLM performance may not be very significant. We will demonstrate this point through the results of ablation experiments. Consequently, we have elected to utilize the Yi-34B model, which has a moderately sized parameter volume, alongside the Meta-Llama-3-8B model, which features a smaller parameter count.
\paragraph{Yi-34B}
The Yi-34B model\citep{Young2024YiOF} is a middle-scale language model trained on a diverse corpus of 3 terabytes, designed with a bilingual focus. It has demonstrated significant potential across various aspects such as language comprehension, common sense reasoning, and reading comprehension, indicative of the emergent capabilities associated with large models\citep{Wang2023UnleashingTE}. This emergent capability implies that LLMs can exhibit human-like logical thinking, reasoning, and analytical abilities, which are precisely the skills that LLMs need to engage in reasoning with the aid of prompts. Suitable for a multitude of applications, the Yi-34B model is fully accessible for academic research and concurrently offers free commercial licensing upon application.
\paragraph{Meta-Llama-3-8B}
The Llama-3 model represents an auto-regressive language model engineered with an enhanced Transformer \citep{Vaswani2017AttentionIA} architecture. It features refined iterations that employ supervised fine-tuning (SFT) coupled with reinforcement learning augmented by human feedback (RLHF), ensuring alignment with human-centric values of utility and safety. Llama 3 has been pre-trained on an extensive corpus exceeding 15 trillion tokens, sourced from a variety of public repositories. For fine-tuning, it leverages a compilation of publicly accessible instructional datasets alongside a substantial collection of over 10 million examples annotated by humans.
\subsection{Dataset Introduction}
The BIG-Bench Hard (BBH) is a subset culled from the original BIG-Bench assessment suite, focusing on tasks that pose a challenge to existing language models. BBH comprises 23 tasks and 27 sub-datasets, and when creating the BBH dataset, researchers adhered to specific filtering criteria, including the number of examples in the task, the presence of human rater performance data, the type of task, and the performance of previous models, among others. This dataset is designed to drive improvements in language model performance on complex reasoning tasks and provides a valuable benchmark for future research. This dataset is relatively difficult and covers a wide range of topics, with many of its sub-datasets focusing on assessing the reasoning capabilities of LLMs, making it an ideal tool to test the abilities of PRS.
\subsection{Set Up}
We have confirmed the linear relationship between the complexity of a problem and the number of steps required to solve it through experiments conducted across multiple datasets. On the Llama3 model and the BBH dataset, this linear relationship is specifically manifested as the number of steps required to solve a problem being half of its complexity. Therefore, once we have ascertained the complexity level of a problem via a Large Language Model (LLM), we utilize this linear relationship to translate the complexity into the corresponding steps necessary for problem resolution.
\subsection{ Evaluation Metrics.}
After obtaining the answer from the Large Language Model (LLM), it is necessary to compare it with the correct answer to assess its accuracy and calculate the rate of correctness. There are two primary methods for comparison: The first method involves the LLM itself comparing the provided answer to the correct one and then making a judgment. This approach is contingent on the performance of the LLM, which carries the potential for error. The second method is applicable only when the correct answers in the dataset adhere to a more uniform and standardized format. In this case, we simply instruct the LLM to pay attention to the format when presenting the answer. Subsequently, we can employ regular expressions to extract the answer and use string methods for comparison. This method demands less from the LLM's performance but has a more limited range of applicability. The answer format for the problems in the BBH dataset is relatively standardized, therefore we can employ the second method, which is more accurate for making judgments.

Since the correct answers in the dataset we used have specific formatting requirements, we provided the LLM with the answer to the first sample in the dataset as a template, instructing the LLM to mimic the format of the first answer when responding. The evaluation of the LLM's responses begins with the second sample, ensuring that the format of the LLM's answers is maintained while preventing data leakage. 

Before determining the correctness of the answers, we categorized the types of answers in the dataset, dividing them into single-choice, multiple-choice, numerical, etc., and formulated appropriate regular expressions to match the answers accordingly.
\subsection{Result}
We conducted extensive experiments on the BBH dataset, where out of its 27 sub-datasets, only 24 were found to be valuable for exploration. The "dyck\_languages" and "multistep\_arithmetic\_two" tasks, which were discarded, primarily assess the LLM's ability to discern symbols, which is unrelated to the performance of the prompt. PRS outperformed CoT in 19 out of 25 sub-datasets, the accuracy rate improved from 36\% with CoT to 44\% with PRS, which is an increase of 22\%.
\subsection{Ablation Experiment}
By applying this framework to the Yi-34B model, we also compared the CoT and PRS methods and found that the average accuracy of PRS increased from 45\% with the CoT method to 49\%, achieving a 9\% improvement. This demonstrates that our method is also applicable to LLMs with a larger number of parameters and also validates our viewpoint that the enhancement effect of prompt design methods on LLMs with large parameter volumes is relatively less pronounced.
\begin{figure}[h]
\centering
\includegraphics[width=0.8\linewidth]{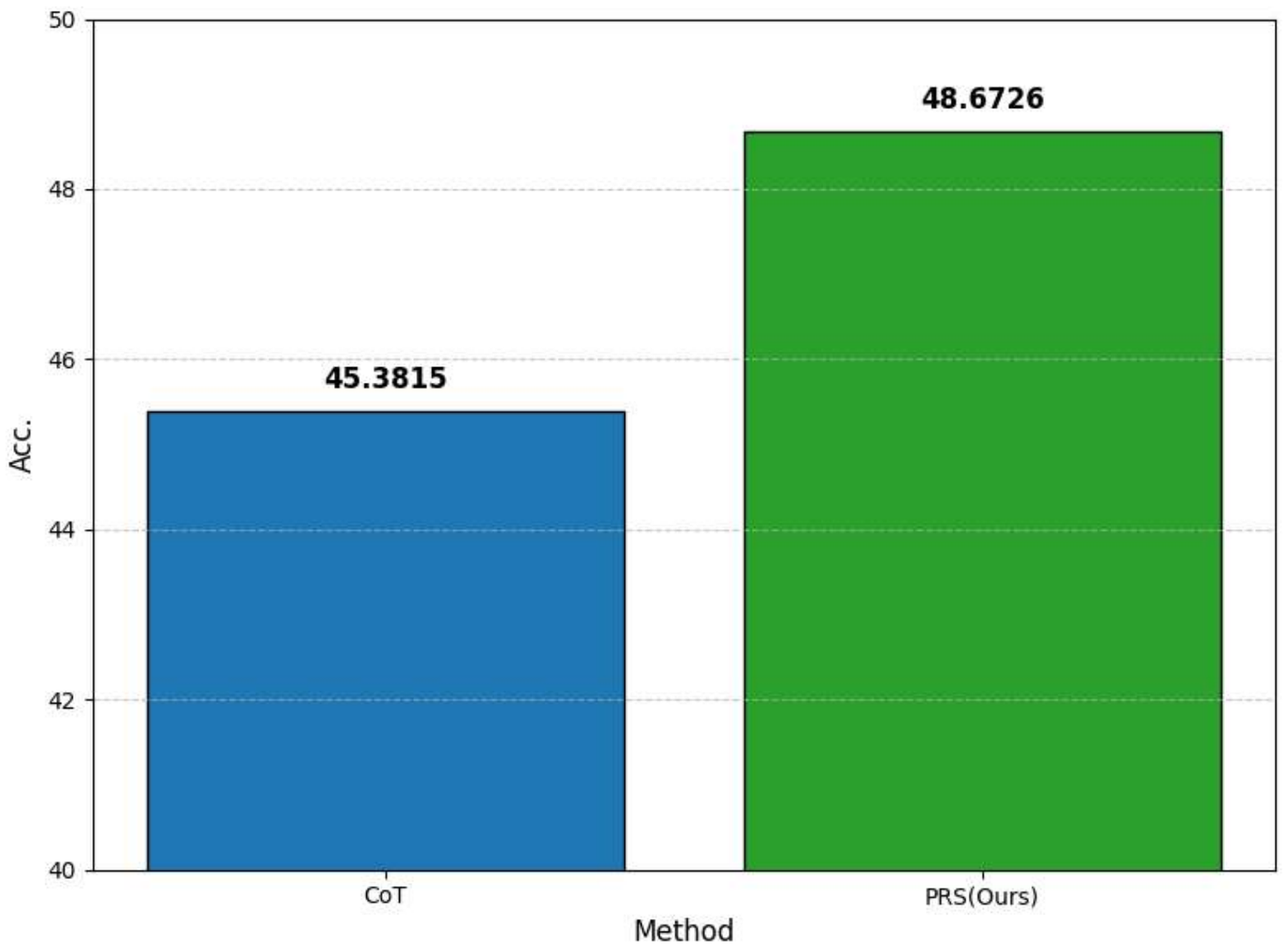}
\caption{The ablation experiments based on the \textbf{Yi-34B} model on the BBH dataset. The results represent metrics on the formal\_fallacies subset.}
\vspace{20 pt}
\label{fig:ablation}
\end{figure}
\section{Conclusion}
\label{sec:Conclusion}

Traditional prompt design methods can be divided into LDP and EDP, where EDP heavily relies on the subjective experience of human experts. Its immutable characteristics once involved make it waste redundant computational resources when dealing with simple problems. On the other hand, LDP is generated and optimized by LLMs, but due to the lack of effective supervision during the prompt generation process, it is prone to error accumulation. We have integrated the advantages of both by designing a brand-new prompt framework called PRS. It eliminates the high dependence on human expert knowledge through automatic prompt design and saves computational resources. By effectively supervising the complexity of the problem, it prevents error accumulation during the reasoning process. We have demonstrated the effectiveness of this framework by comparing its performance with the CoT method using the Llama3-7B model across multiple domain datasets, as well as with the ablation experiments using the Yi-34 model.
\section{Limitations}
By summarizing traditional prompt design techniques, we have categorized traditional prompt design methods into EDP (Expert-Designed Prompts) and LDP (LLM-Derived Prompts). Upon discovering the complementary nature of the advantages between the two, inspired by the differentiation phenomenon of stem cells in biological principles, we have innovatively proposed a new prompt structure: Prompt Recursive Search (PRS). This structure can automatically design prompts for problems, which has the advantage of saving computational resources. At the same time, by judging the complexity of the problem, it supervises the errors in the prompt design process, thereby ensuring accuracy. Through experiments in multiple domains and with various models, we have demonstrated the effectiveness of PRS. However, it is undeniable that our framework still has certain shortcomings, which are as follows:
\begin{enumerate}
    \item We mandate that the Large Language Model (LLM) outputs answers in a specific format through our prompt, as deviation from this format would render the true-false judgment infeasible. However, the LLM is not always compliant with the prescribed output format, and answers that do not conform to the format could be either erroneous or correct. Our condition for deeming an LLM-provided answer as correct stipulates that not only must the content be accurate, but the format must also be correct. This essentially elevates the criteria for acceptable responses. Situations where the content is correct but the format is not are not accounted for in our assessment, leading to an underestimation of our true accuracy rate in the statistics we compile.
    \item Our framework is predicated on a crucial assumption: the complexity of a problem is directly proportional to the number of steps required to solve it. We have substantiated this intuitive assumption through numerous experiments, yet it is important to recognize that this assumption represents a macroscopic rule and there are a few exceptional cases that do not conform to it. Consequently, the number of steps necessary to resolve a problem remains a topic worthy of exploration. We thus leave this question for future research endeavors.
    \item Even for the same input, the responses from the Large Language Model (LLM) can vary with each invocation, thus reflecting that the performance of the Prompt Recursive Search (PRS) framework has an inherent degree of randomness. The experimental results presented in this paper are the average of three trials, which may still differ from the true capabilities of the method.
    \item Due to the inherently strong capabilities of Large Language Models (LLMs) with a larger number of parameters, applying the Prompt Recursive Search (PRS) method proposed in this paper does not yield a significant enhancement in this kind LLM's abilities. In fact, the effectiveness of this framework is contingent upon the LLM's analytical capacity regarding the problem at hand. Consequently, LLMs with a larger parameter count are better suited to leverage the full potential of the PRS framework. Our ablation study proved this perspective but did not delve into a detailed investigation. We reserve the exploration of this aspect for future work.
\end{enumerate}
In summary, we have proposed a brand-new method for prompt design and have validated it through extensive experiments, providing a starting point for future research.








\begin{ack}
By using the \texttt{ack} environment to insert your (optional) 
acknowledgements, you can ensure that the text is suppressed whenever 
you use the \texttt{doubleblind} option. In the final version, 
acknowledgements may be included on the extra page intended for references.
\end{ack}


\clearpage
\bibliography{mybibfile}

\end{document}